# Artificial Intelligence-Based Methods for Precision Medicine: Diabetes Risk Prediction

Farida Mohsen[1], Hamada R. H. Al-Absi[1], Noha A. Yousri[2], Nady El Hajj[1,3] and Zubair Shah[1*]

[1]College of Science and Engineering, Hamad Bin Khalifa University, Qatar Foundation, 34110, Doha, Qatar.
[2]Genetic Medicine, Weill Cornell Medicine-Qatar, Qatar Foundation, Doha, Qatar.
[3]College of Health and Life Sciences, Hamad Bin Khalifa University, Qatar Foundation, 34110, Doha, Qatar.

*Corresponding author(s). E-mail(s): zshah@hbku.edu.qa;

**Abstract**

The increasing prevalence of type 2 diabetes mellitus (T2DM) and its associated health complications make developing predictive models for T2DM risk prediction crucial for implementing precision medicine and prevention strategies. Artificial intelligence (AI) predictive models have recently been widely developed for T2DM risk prediction; however, a comprehensive review of the advancements and challenges of AI-based prediction models for T2DM risk prediction is currently lacking. To this end, this scoping review aims to comprehensively synthesize and analyze the existing literature on the state of AI-based models for T2DM risk prediction. A systematic search of longitudinal studies was performed across four databases, including PubMed, Scopus, IEEE-Xplore, and Google scholar. The PRISMA-ScR (Preferred Reporting Items for Systematic Reviews and Meta-Analyses Extension for Scoping Reviews) guidelines were followed. Forty studies that met our inclusion criteria were included in the review, with the majority published in the past four years. The results showed that traditional machine learning (ML) models were more widely used than deep learning (DL) models. Electronic health records (EHR) were the most commonly used data modality, followed by multiomics, while medical imaging was the least used. Most studies employed unimodal AI models (n=30) that relied mostly on EHR data, while only ten used multimodal models combining different data modalities. The results demonstrated promising performance of both unimodal and multimodal AI models in T2DM risk prediction, with multimodal models outperforming unimodal models. All studies conducted internal validation, while only five conducted external validation. Most studies reported discrimination measures using the area under the receiver operating curve (AUC) and varying degrees of classification metrics. However, only five studies reported on the calibration of their models. Interpretability methods that identify and explain important risk factors and biomarkers used by the prediction models were reported in only half of the studies. The review revealed a limited number of studies that reported novel biomarkers, with most focusing on commonly known risk factors. Furthermore, the review found that a limited number of studies made their methods available to the scientific community using open-source code. Overall, our review provides valuable insights into the current state and limitations of AI-based models for T2DM risk prediction and highlights the challenges associated with their development and clinical implementation.

# 1 Introduction

Precision diabetes medicine represents a cutting-edge approach to diagnosing, predicting, and treating diabetes, one that considers individual variations and integrates diverse data sources to provide a comprehensive understanding of the individual's health status, predisposition, and treatment response [1, 2]. T2DM is the most widespread form of diabetes, and its global incidence and prevalence are growing, putting a significant burden on healthcare systems. Given the economic and personal impact of T2DM, including decreased productivity, higher healthcare costs, severe complications, and shortened lifespan, there is a pressing need for preventive efforts.

Precision prognostics, a critical aspect of precision diabetes medicine, aims to develop predictive models to estimate an individual's risk of developing T2DM and its complications based on their risk profiles [1]. This enables the identification of high-risk individuals, allowing for personalized prevention strategies and targeted treatments to delay or prevent the onset of the disease and its complications [1, 3]. The American Diabetes Association (ADA) and the European Association for the Study of Diabetes (EASD) Consensus Report support this approach and recommend targeting high-risk individuals with lifestyle interventions and glucose-lowering medications to prevent or delay the onset of T2DM [1].

For prognostic models to be implemented into routine care, they must go through different stages, including model development, evaluation, and translation to clinical decision support [4–6]. The development of these models entails utilizing longitudinal data on individuals' biological characteristics, lifestyle, and environment [7]. The next essential step is evaluating the prognostic model's performance. An effective prognostic model is characterized by its ability to accurately estimate an individual's risk—that is, predictions made by the model closely match what is observed (calibration), its ability to reliably differentiate between individuals who are at high risk and those who are at low risk of developing the condition (discrimination), and its ability to perform well in different populations (generalizability). The validation of a prognostic model involves testing its calibration and discrimination through either internal or external means, with external validation being preferred [7, 8].

Many researchers in the literature have proposed T2DM risk prediction models that usually involve risk scores [9, 10]. However, these models have limitations, such as the use of a limited number of risk factors as input features that do not consider the complex interplay of different biological systems involved in the development of T2D [10, 11], and potential bias caused by dependence on prior publications for predictor identification [12]. Artificial intelligence (AI), particularly machine learning (ML) and deep learning (DL), can overcome these weaknesses and has gained increasing attention in medical research for its potential to analyze large biomedical datasets, including electronic health records (EHRs), medical imaging, multi-omics data, behavioral/wellness, and environmental data [13, 14]. In recent years, ML algorithms have emerged as a promising tool for developing predictive models for T2DM by analyzing complex and multidimensional datasets to identify high-risk individuals, identify patterns and biomarkers associated with T2DM development, and guide personalized interventions for prevention. While most existing AIbased T2DM predictive models are focused on one data modality (i.e., EHR), there has been a recent shift towards multimodal models that integrate diverse data sources for a holistic view of the individual and more precise prevention.

As the landscape of AI-based T2DM predictive models is rapidly evolving, a comprehensive overview of the field's current state is needed. To this end, our scoping review aims to synthesize and provide a comprehensive analysis of the use of AI-based algorithms for T2DM risk prediction. This involves examining the various AI models utilized, the types of data and predictors employed, the datasets and evaluation metrics used, as well as the most significant risk markers reported that can guide preventive and early intervention strategies. Additionally, we critically evaluate the limitations and challenges of AI models in this context and highlight the challenges associated with their clinical implementation. Our review also aims to identify the knowledge gaps in the field, calling attention to the areas where further research is needed to advance the application of AI in T2DM risk prediction. To our knowledge, few reviews in the literature have explored the application of AI techniques in diabetes. However, these reviews differ from ours in terms of scope and coverage. For instance, some earlier works focused solely on evaluating AI models for diabetes detection [15, 16] or predicting diabetes complications [17]. One study conducted a meta-analysis of the predictive ability of ML models for T2DM risk prediction using a limited sample size of 12 studies [10]. In contrast, our scoping review focuses on studies that utilize AI methods for T2DM risk prediction, focusing on those that leverage longitudinal data to construct predictive models. Our review encompasses a wide range of AI models for T2DM risk prediction, including unimodal and multimodal models, and covers a total of 40 studies that use various data modalities, such as EHRs, multi-omics, and imaging data. Table 1 highlights the main differences between our scoping review and previous reviews in the field, emphasizing the need for a new and updated review in this area.

**Table 1:** Literature review comparison

| Previous reviews | Year | Scope and coverage | comparative contribution of our review |
|---|---|---|---|
| Machine learning and artificial intelligence based Diabetes Mellitus detection and self-management: A systematic review [15]. | 2022 | Their review focused on the performance of AI models in diabetes detection. | Our review focused on AI models in T2DM risk prediction, specifically considering studies that utilized longitudinal data in their predictions. |
| Use and performance of machine learning models for type 2 diabetes prediction in community settings: A systematic review and meta-analysis [16]. | 2020 | They performed a systematic review to prove whether ML exhibited discrimination ability to predict and diagnose T2DM; their review included only eight studies that performed T2DM risk prediction. | Our review focused on AI models in T2DM risk prediction, specifically considering studies that utilized longitudinal data in their predictions. Moreover, our review covered more studies with a wider range of AI models for T2DM prognosis, including unimodal and multimodal models. |
| Microvascular complications in type-2 diabetes: a review of statistical techniques and machine learning models [17] | 2020 | They conducted a review on microvascular complications in diabetes (retinopathy, neuropathy, nephropathy) | Our review focused on AI models in T2DM prognosis rather than its complications. |
| Predictive ability of current machine learning algorithms for type 2 diabetes mellitus: A meta-analysis [10] | 2022 | They conducted a review with meta-analysis to evaluate the current ability of ML algorithms for T2DM prognosis retrieving 12 studies till 2020. All the retrieved studies used EHR data. | Our review focused on AI models in T2DM risk prediction, specifically considering studies that utilized longitudinal data in their predictions. Moreover, our review covered more studies (n=40) with a wider range of data modalities including EHR, multi-omics, and imaging. |

# 2 Methods

In conducting this scoping review, we adhered to the guidelines as recommended by PRISMA-ScR [18].

## 2.1 Search strategy

A systematic search was conducted in four databases, including Scopus, PubMed, IEEE Xplore, and Google Scholar. It is worth mentioning that MEDLINE is covered in PubMed. Only the first 100 relevant studies from Google Scholar were considered for the review, as search results beyond this number rapidly lost relevance and were not pertinent to the scoping review topic. In addition to the database search, reference lists of the included studies were screened to identify additional relevant literature. Search strings were established via literature searches and domain expertise.

In this scoping review, we focused on studies that used AI methods for T2DM prognosis using longitudinal datasets. Thus, the search string consisted of four major terms connected by AND operator: [ ("Artificial Intelligence" OR "machine learning" OR "deep learning") AND ("prediction" OR "prognosis") AND ("diabetes" OR "T2DM") AND ("longitudinal")]. We used different forms of each term and adapted each string for each database. The complete string can be found in Appendix 1 of the supplementary material. Gray literature was not included in this scoping review in an attempt to only include peer-reviewed studies.

## 2.2 Inclusion and exclusion criteria

The authorship team jointly developed the selection criterion. To be eligible for inclusion in the review, studies had to meet the following criteria: (1) the study used longitudinal data; (2) the primary aim of the study was to use AI/ML algorithms to predict the future development of T2DM; For our context, AI models meant to be classical ML models, DL models, ensemble learning, etc as mentioned in the search terms in Appendix 1 of the supplementary material; (4) the study was conducted using humans; (5) the study used any medical data modality including imaging, EHR, multi-omics; we did not limit our study to one or two medical data modalities, and we considered the studies that fused different data sources of the same type as multimodal; for example, a study using genomics with metabolomics was considered as multimodal under our premise; (6) only original research, peer-reviewed studies and conference proceedings were included and limited to the English language.

We excluded studies that used classical statistical models such as regression analysis and those that performed T2DM classification or diagnosis using cross-sectional data or studies related to type 1 diabetes, gestational diabetes, or T2DM-related complications and their severity assessment. Additionally, studies that used non-human derived data, non-English publications, review articles, conference abstracts, proposals, editorials, commentaries, letters to editors, preprints, and short letter articles were excluded. All papers underwent a 2-person verification for inclusion/exclusion in/from the manuscript.

## 2.3 Study selection and data extraction

We utilized the Rayyan web-based review management tool [19] for screening and study selection process. One reviewer (F.M.) conducted the literature search. After removing duplicates, we screened the studies based on their title and abstract to identify potentially eligible studies. A full-text screening of all potentially included studies was conducted to extract data specific to each study. The study selection and information extraction were done by two reviewers (F.M. and H.R.A.). Discrepancies were addressed through discussion, and in cases where a mutual agreement could not be established, a third author (Z.S.) was consulted for resolution.

An evidence form was created and tested on five studies to establish a systematic and precise data extraction method. Data extracted from the studies included the titles, first author's name, publication year, publication venue, country of the first author's institution, the study objective, the study design, follow-up period, and data source (public or private). Additionally, we recorded the number of different data modalities and their categories, such as EHR, EHR/images, or EHR/multi-omics/images. For instance, a study that used clinical measures (structured EHR) and retina images as inputs for the AI was categorized as the "Imaging/EHR" subtype. We also extracted the type of data used in each modality, such as the type of imaging or multi-omics data.

For model development, we extracted information on the type of AI utilized (such as ML, DL, or a combination of both), the specific ML algorithm applied, the data fusion technique utilized for multimodal models, the methods applied to address data imbalance and missing data, if any, and the code availability. Additionally, we recorded the validation type (internal or external), the nature of the internal validation, and the evaluation measures used. We further extracted information about the interpretability methods used to determine feature importance, if any, and the most significant risk factors and biomarkers reported by the studies.

## 2.4 Data synthesis

Following data extraction, we used a narrative approach to synthesize the data. We aggregated information collated in the initial evidence form to determine the main themes of AI's use in T2DM future risk prediction in the currently published literature. Given the heterogeneity of data modalities, AI methodology, implementation details, datasets, and evaluation measures, we analyzed the studies from different perspectives. Firstly, we examined the type and combination of data modalities used in each study, including EHRs, medical imaging data, multi-omics data, and their combinations. Secondly, we categorized the type of AI models based on the number of data modalities used as input for these models, either unimodal or multimodal. We then investigated the AI models used in each study in both categories and the fusion strategy used to integrate data for multimodal models. We also discussed the implementation details of the models, such as data imbalance treatment, missing values treatment, and interpretability methods. Thirdly, we analyzed the type of reported risk factors and biomarkers in each study, identifying the specific clinical or demographic features used to predict T2DM risk and any biomarkers or genetic markers. Fourthly, we evaluated each study's validation approach and evaluation measures. Lastly, we assessed the data sources and their availability in each study, including public or private data sources and open or restricted access data. Examining the studies from multiple perspectives gave us a comprehensive understanding of the current literature on using AI for T2DM risk prediction.

## 3  Results

### 3.1  Search results

After an initial screening based on titles and abstracts, 64 studies were considered eligible for full-text screening. 31 studies were excluded after a thorough full-text review against established inclusion criteria for various reasons.Additionally, seven studies were identified through forward and backward reference screening, leading to a final of 40 studies that underwent data extraction and synthesis. The overall study screening and selection process is depicted in Fig. 1a.

### 3.2  Demographics of the included studies

Among the included studies, 37 were peer-reviewed journal articles, and 3 were conference publications. The studies were published between 2012 and 2022, with the majority being published in the past four years (n=29). This suggests a growing interest in using AI models for T2DM prognosis, which can be attributed to advancements in AI models and the increasing focus on precision

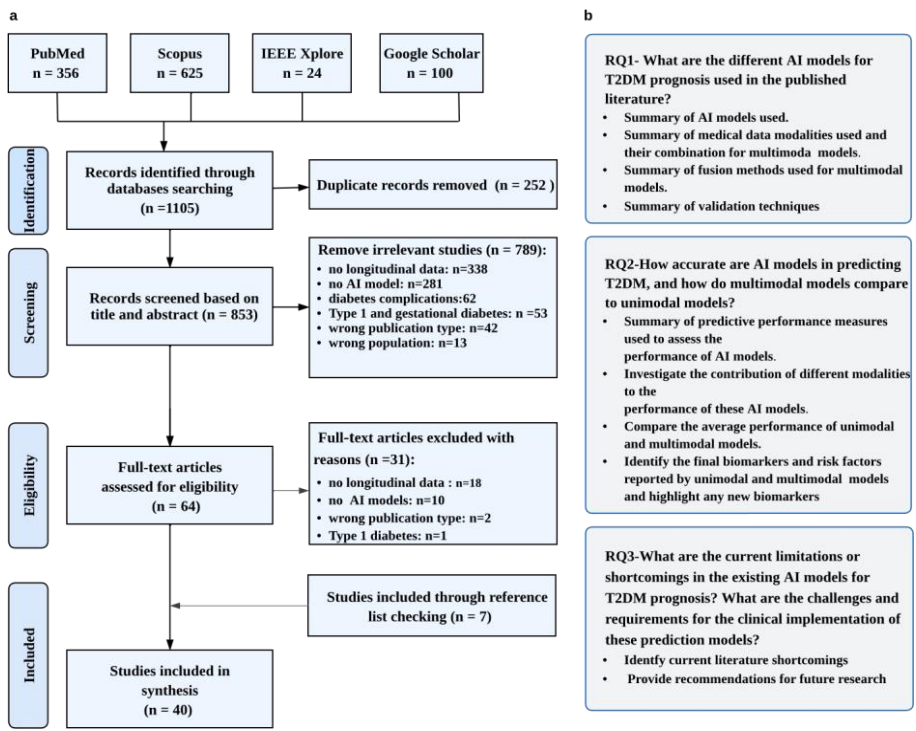

**Fig. 1** a: PRISMA for identifying, screening, and selecting studies for synthesis; b: Research questions posed.

medicine. The studies were published in a diverse range of countries, with the majority being from the United States (n=6), followed by China (n=7) and Korea (n=4). The diversity in the country of publication highlights the global interest in using AI models for T2DM prognosis. The distribution of studies by publication venue type, year, and country is presented in Fig.2

## 3.3 Study designs and prediction horizons

All of the AI prediction models included in this scoping review were developed using data from longitudinal cohort studies in accordance with the inclusion criteria outlined in this study. The study designs varied among the included studies, with the majority being retrospective cohort studies (n=18) [11, 20–36] that utilized data from large clinical databases and registries. Other study designs included prospective cohort studies (n=10) [37–46], case-control studies (n=11) [47–57], and one case-cohort study [58]. Detailed information can be found in Appendix 2 in the supplementary material.

In terms of prediction horizon, the studies reported a wide range of horizons, with most of the studies focusing on medium-term predictions (n=18) [20, 20, 22–24, 26, 31, 33, 39–41, 44–47, 51, 52, 58], such as predicting diabetes onset within 5-10 years. Additionally, 13 studies [24, 28–30, 34, 36–38, 50, 54–57] focused on short-term predictions, such as predicting diabetes onset within less than five years, and eight studies predicted T2DM onset within five years [22, 24, 25, 27, 32, 35, 38, 47]. A smaller number of studies (n=5) [21, 42, 43, 48, 49] reported long-term predictions, such as predicting T2DM onset within more than ten years. It is worth noting that some of the included studies reported multiple prediction horizons, so the sum of the number of studies in each category may not be equal to the total number of included studies.

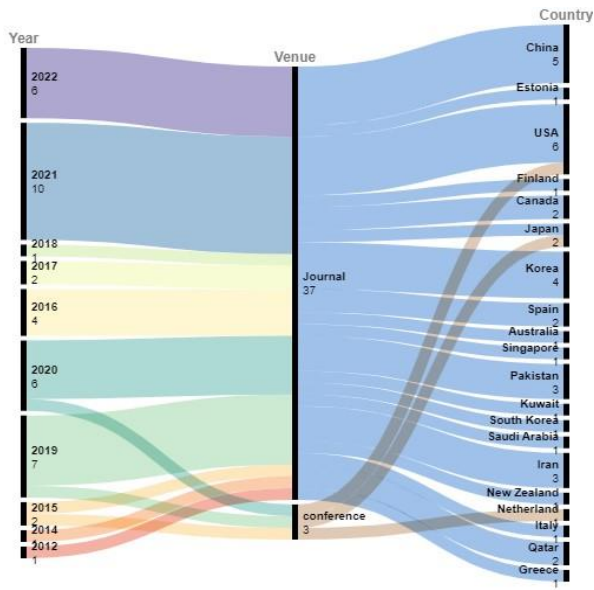

**Fig. 2** The studies' distribution by the type of publication venue, year, and country.

## 3.4 Data modalities

The data modalities used in the included studies varied widely, reflecting the diversity of information that can be used to predict T2DM. EHRs were used by all 40 studies and included information such as sociodemographic characteristics, family history of diabetes (FHD) and other diseases, lifestyle risk factors, anthropometric measures, glycemic traits, blood lipid, blood pressure factors, etc. Other data sources utilized included multi-omics data, such as single nucleotide polymorphisms (SNPs) associated with T2DM risk, metabolomic data in the form of metabolite levels in the blood, and microbiome data. Medical imaging data was also used, including information from various modalities such as computed tomography (CT) and fundus images. It is worth noting that several of the studies included in this scoping review fused different data modalities to constitute multimodal predictive models. The most common combination was EHR and multi-omics data, while some studies combined medical imaging with EHR data. Table 2 summarizes the data modalities and their combinations used in the studies.

Table 2 Data types used by the included studies.

| Data type | Number of studies | Study reference |
|---|---|---|
| EHR | n=40 | [11, 19–57] |
| EHR+ genetics | n=2 | [37, 49] |
| EHR+ metabolomics | n=2 | [26, 50] |
| EHR+ metabolomics + genetics | n=2 | [46, 48] |
| EHR+ Microbiome | n=1 | [38] |
| EHR + Fundus images | n=2 | [35, 45] |
| EHR + CT | n=2 | [36, 58] |

## 3.5 T2DM AI-based risk prediction models

The type and number of data modalities used to train a prediction model can significantly affect the model's performance and impact the model's reliability and prediction outcomes. With this in mind, we categorized the AI models developed in the reviewed studies into unimodal and multimodal categories. Unimodal models use a single type of data as input, whereas multimodal models incorporate multiple data sources as input.

### 3.5.1 Unimodal predictive models

In the scoping review, 30 studies that focused on the use of unimodal models for T2DM risk prediction were identified and accounted for the majority of the included studies [11, 20–25, 27–34, 39–44, 47, 51–58]. A summary of their characteristics can be found in Table 1 of Appendix 3 in the supplementary information. A variety of AI algorithms were utilized in the development of these predictive models, with classical ML being the most frequently employed compared to DL models. The classical ML models employed were very diverse. Among them, decision trees (DT) and their variants such as classification and regression tree (CART), quick unbiased efficient statistical tree (QUEST), commercial version (C5.0), and DT using the CHAID method were the most widely used, with ten studies implementing them. Moreover, linear regression (LR) was used in ten studies. Random forest (RF) was used in nine studies, while support vector machines (SVM) were implemented in eight studies. Naïve Bays (NB) classifiers were implemented in five studies, while K-nearest neighbor (KNN) and extreme gradient boosting (XGBoost) were utilized in four studies each. Ensemble learning, which involves combining the predictions of multiple models to improve the overall accuracy of the prediction, was used in four studies employing different forms of voting, such as soft voting and weighted voting. Hidden markov models (HMM) was used in three studies, while gradient boosting machine (GBM) and linear and quadratic discriminant analysis (LDA and QDA) classifiers were used in two studies each. The least used algorithms were K-means, AdaBoost, cox regression, and multiple instances learning boosting (MIL-Boost) with one study each. Regarding DL algorithms, feed forward neural networks (FFNN) were used in three studies, followed by long short-term memory (LSTM) and probabilistic neural network (PNN) in one study each. Table 4 presents the distribution of unimodal AI models across different studies.

**Table 3** Distribution of Unimodal AI Models in Studies.

| Unimodal AI models | Study references |
| --- | --- |
| Classical ML models | |
|   DT | [21, 27, 28, 41–44, 54–56] |
|   LR | [11, 22, 24, 25, 27–29, 39, 54, 55, 57] |
|   RF | [11, 20, 22, 27, 28, 39, 54, 55, 57] |
|   SVM | [21, 24, 39, 51–54, 57] |
|   NB | [27, 42, 52, 54, 55] |
|   KNN | [21, 24, 25, 39] |
|   Ensemble models | [27, 52, 55, 57] |
|   XGBoost | [28, 32, 34, 57] |
|   HMM | [23, 33, 59] |
|   GBM | [22, 39] |
|   LDA and QDA | [21, 25] |
|   K-means | [58] |
|   MIL-Boost | [31] |
|   AdaBoost | [52] |
| DL models | |
|   FFNN | [22, 30, 39] |
|   LSTM | [47] |
|   PNN | [42] |

Around half of the unimodal studies compared several ML algorithms and then selected the best-performing one for prediction. In studies that compared two or more algorithms, DT was found to have the best performance in five studies [21, 41, 43, 44, 56]. Meanwhile, ensemble models showed the best performance in four studies [27, 52, 55, 57]. Both RF [11, 22, 54], XGBoost [28, 32, 34], and HMM [23, 33, 40] each had the best performance in three studies. The performance of the algorithms was measured using AUC metric in most of the studies (n=23). The best-selected models in the unimodal studies had AUC values ranging from 0.74 to 0.92, with three studies having values below 0.74 and only one exceeding 0.92. However, it is crucial to note that these results are based on individual studies and are not directly comparable due to the use of different datasets and risk predictors, evaluation metrics, and follow-up periods across studies.

All unimodal predictive models used EHR data as input except one study used imaging [58]. The EHR risk factors and biomarkers used as inputs by the included studies can be broadly categorized into several categories: (1) sociodemographic and family history of diabetes (FHD); (2) lifestyle; (3)

anthropometric measures of body size and composition; (4) glycemic traits that include measures of glucose control; (5) blood lipid and blood pressure factors that include measures of cholesterol, triglycerides, and blood pressure; (6) inflammatory biomarkers include measures of inflammation, such as C-reactive protein-to-albumin ratio (CAR); (6) other biomarkers: this includes measures of liver function, such as liver enzyme levels, or measures of adiposity, such as circulating adiponectin levels; (7) medications and different diseases history. Table 1 in Appendix 3 of the supplementary information includes the different EHR categories and features among studies. Only one study [58] in this category used imaging data represented in CT scans to investigate the relationship between different fat distribution patterns and the risk of developing T2DM.

Overall, the results showed that unimodal predictive models had moderate to high performance in predicting T2DM risk, with an average AUC of 0.809. However, it is essential to note that it is difficult to determine exactly which AI model is performing the best, as the type and combination of input risk factors can significantly influence the model's performance. For example, the XGBoost algorithm was used in three studies [28, 32, 34], and the reported AUC was 0.91,0.826,0.67, respectively. Each of these studies used different risk factors, causing the algorithm to perform well with certain combinations of risk factors but not as well with others. Additionally, the results show that the prediction horizon also impacts the performance of the same model. In studies that performed prediction over different periods showed that the discriminatory power decreases as the prediction horizon increases [22, 24, 47, 54].

### 3.5.2 Multimodal predictive models

Surprisingly, multimodal predictive models were less common in the studies included in this review. A total of ten studies used multimodal predictive models [26, 35–38, 45, 46, 48–50], as shown in Table 2 of Appendix 3 in the supplementary information. These models employed various fusion strategies, including early fusion and joint fusion, as defined in [60, 61].

**Early fusion**

Early fusion models combine multiple data modalities at the data level before feeding the features into the AI predictive model. This allows the model to take advantage of the complementary information provided by different data sources. A total of seven studies examined the use of early fusion models for T2DM risk prediction incorporating multiple data modalities such as EHR and multi-omics [26, 37, 38, 46, 48–50]. For example, two studies combined genetics and EHR features [37, 49], while two other studies fused metabolomics data with EHR [26, 50]. Additionally, two studies integrated genomics, metabolomics and clinical data [46, 48], and one incorporated microbiome data with EHR [38].

Five of these studies applied feature selection techniques to eliminate extraneous and redundant variables to address the issue of high dimensionality before combining the features from the different modalities [37, 46, 48–50]. These techniques included cox proportional hazards (CPH) [37], J48 decision tree (DT) method [50], random forest feature selection [48], Boruta algorithm [46], and L1 penalized logistic regression ref48. The selected features from the various modalities were then concatenated and fed as inputs into a range of AI models for T2DM prognosis. Classical ML algorithms were the most commonly used, with RF being the most popular, employed in four studies [37, 38, 46, 48]. Other methods such as GBM [37] NB [50], Regularized Least Squares (RLS) [26], and J48 DT [50] were used only in one study each. Deep neural networks (DNN) were also used in two studies [37, 49].

Two studies investigated the value of multiple genetic variants in T2DM prediction [37, 49]. These studies included in this review used genetic risk factors in combination with classical risk factors to improve the performance of their AI prediction models. In [37], the findings indicated that prediction by clinical risk factors was significantly improved when genetic information was added. Another study by Kim et al. [49] compared performance of LR and DNN models using different numbers of SNPs without and with clinical factors. They found that both models contain only 96 and 214 SNPs, had poor discriminative ability and did not outperform a clinical model with classical risk factors. However, when more SNPs were included (399 and 678 SNPs), both LR and DNN models showed better AUC than the clinical model. Importantly, the combination of clinical factors had significantly improved AUC of DNN models across the board. The incorporation of clinical factors had a more pronounced effect when fewer SNPs were included. Overall, the results of this study suggest that combining genetic risk factors with classical risk factors may improve the performance of AI prediction models for T2DM prognosis; this is especially true when a larger number of genetic variants are included in the model. These findings highlight the potential value of incorporating genetic information into T2DM prediction models for more accurate and personalized risk assessment.

Two studies [26, 50] have also explored metabolomic biomarkers as potential risk predictors for T2DM. In one study [26], the use of the entire metabolome as a predictor of T2DM resulted in significantly improved performance compared to a model based on clinical risk factors alone, with an AUC of 0.77 versus 0.68, respectively. Moreover, a subset of metabolite signatures achieved an average AUC of 0.75, showing higher predictive performance than the reference clinical-only model (AUC = 0.68). However, combining the panel of selected metabolite features with clinical variables achieved the highest predictive performance, resulting in an improved AUC of 0.78. This combination outperformed both the clinical and metabolite-only models, demonstrating the significance of combining both types of data for improved predictive accuracy. Another study [50] reported identifying 21 metabolites that significantly differ by incident T2DM status resulting in an AUC of 0.769. Moreover, they combined the metabolites data and glucose risk factors achieving an AUC of 0.754. This suggests that, in this particular study, metabolomics risk factors alone were more effective at predicting T2DM than when combined with classical risk factors. In conclusion, using metabolomics risk factors is effective at predicting T2DM in studies.

In two studies [46, 48], genomics and metabolomics risk factors in combination with classical risk factors was found to improve performance of AI prediction models for T2DM. One study [46] reported that incorporating both genomics and metabolomics biomarkers with classical risk factors improved the model's performance significantly, with an AUC of 0.8835 when using three data types compared to an AUC of 0.876 when using genetic and classical risk factors or 0.84 when using only classical risk factors. Similarly, another study [48] also found that the combination of genomics, metabolomics, and clinical risk factors had the best performance for predicting T2DM with an AUC of 0.960 compared to an AUC of 0.586 for the genomics-only model and 0.798 for the clinical risk factors-only model. In summary, the use of genomics and metabolomics in combination with classical risk factors has shown promising results in the prediction of T2DM. The incorporation of these biomarkers has consistently resulted in improved performance of AI prediction models, potentially due to the increased amount of information available to the model.

Oliver et al. [38] conducted the first longitudinal study assessing the gut microbiome as a predictive factor for various parameters associated with T2DM. Their findings demonstrated that the microbiome, in conjunction with conventional risk factors, can effectively predict multiple metabolic outcomes. Therefore, the prospect of using the microbiome in personalized medicine is promising. However, the true potential of the gut microbiome for predicting T2DM remains unknown.

To assess the potential of multimodal learning through fusion, six studies [26, 37, 38, 46, 48, 49] compared the performance of the fusion models to that of single-modality models. As a result, five of these studies found that early fusion models had improved performance compared to their unimodal counterparts [26, 37, 46, 48, 49]. The average AUC value of the early fusion models was 0.89. These results indicate the potential of early fusion models in improving T2DM risk prediction.

**Joint fusion**

In this scoping review, three studies examined the use of a joint fusion strategy for combining multiple data modalities to develop T2DM predictive models [35, 36, 45]. These studies fused EHR metadata with different types of medical imaging, such as CT and retina scans, as shown in Table 2 of Appendix 3 in the supplementary information. All these studies employed DL models to extract imaging features and jointly learn multimodal feature representation for T2DM prediction.

One study by Zhang et al. aimed to examine the ability of an AI model to predict the future risk of T2DM onset in individuals based on their fundus images and clinical data [35]. To achieve this, they proposed a deep multimodal framework that can better learn complementary features from both image and non-image modalities for T2DM prediction 5 years in advance. They used deep convolutional neural networks (CNNs), specifically a residual network (ResNet50) architecture to convert the fundus image data into a feature vector fusible with other metadata. Then, the image feature vector derived from the CNN model was concatenated with the clinical features of the same patient and fed into a multilayer perception (MLP) for joint learning and prediction. Based on their results, the authors concluded that fusion models of fundus images and clinical data can be used to automate the prediction of T2DM risk in healthy individuals. Similarly, a study by Yun et al. [45] also investigated retinal risk factors with additional traditional risk factors for T2DM screening and prediction using the ResNet18 model. Another study [36] explored the feasibility of combining CT images and clinical data to develop a 1-year risk prediction model for T2DM. Pancreatic CT images were used to extract body composition features, such as abdominal visceral fat volume, subcutaneous fat volume, and pancreas volume, using CNNs. To develop a clinical-image multimodal risk prediction model, the study combined these imaging features with clinical data and fed them into fully connected layers.

The three studies assessed the performance of their joint fusion models against single modality models, either clinical-only or imaging-only. They all demonstrated better fusion performance than their single modality counterparts. The average AUC value of the joint fusion models was 0.86, ranging from 0.844 to 0.89. These results indicate the potential of joint fusion models in improving T2DM risk prediction through integrating multiple data modalities, resulting in more accurate predictions.

## 3.6  Performance comparison of unimodal and multimodal T2DM predictive models

The results of this scoping review demonstrate that AI models based on various data modalities can be effective in predicting the development of T2DM. Unimodal models showed widely varied performance with an average AUC of 0.809. On the other hand, multimodal models demonstrated improved performance compared to their unimodal counterparts, with an average AUC of 0.879, as shown in Fig. 3. However, it should be noted that these results from the included studies cannot be directly compared due to the use of different datasets, evaluation measures, and prediction horizons. The improved performance of multimodal models can be attributed to the additional information that is made available through the fusion of multiple data modalities. By combining various sources of information, multimodal models can capture a complete picture of an individual's T2DM risk, leading to better performance.

While multimodal models show better performance compared to unimodal methods, they also have some limitations and challenges. One of the main drawbacks is their lower scalability, and the time-consuming nature of data concatenation [59]. Different modalities often have different data types, distributions, and scales, making it difficult to combine the data and make predictions effectively. Additionally, multimodal models often require more computational resources, which can be challenging when working with large datasets. Moreover, it's still challenging to understand the relationship between the different modalities in the multimodal frameworks and how they contribute

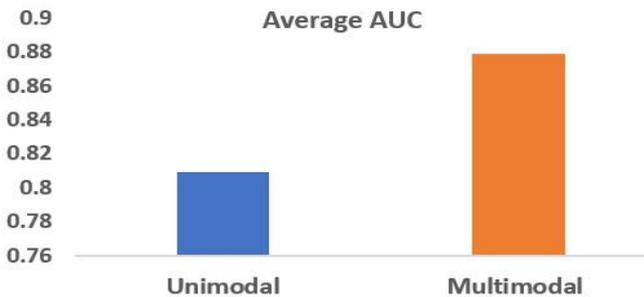

**Fig. 3** performance comparison of unimodal and multimodal AI models for T2DM risk prediction.

to the overall prediction of disease progression. Overall, while multimodal models offer several advantages, they also face limitations in terms of complexity and interpretability. Addressing these limitations is crucial for successfully integrating these models into clinical practice.

## 3.7 Dataset size and class balance

One crucial factor that can affect an AI model's performance is the dataset's size and balance. In the included studies of this scoping review, the size of the datasets used varied greatly, with some studies using datasets containing hundreds of patients and others using datasets containing tens of thousands of patients. The number of participants used to develop the prediction models ranged from 244 patients [50] to 1893901 patients [32]. Detailed information can be found in Appendix 2 in the supplementary material.

In more than three-fourths (n=34) of the included studies in this scoping review, the class distribution within the dataset was imbalanced, meaning that one class was more prevalent than the other [11, 20, 22–25, 27–29, 31, 33– 35, 37–44, 46, 47, 49, 51–53, 55–58]. This can lead to bias in the model. To address this issue, 14 of the studies implemented different techniques. Six of these studies used the Synthetic Minority Oversampling Technique (SMOTE) to generate synthetic samples of the minority (positive) class [23, 27, 30, 39, 41, 42], while two used downsampling [27, 51]. Two studies also partitioned their datasets 100 or more times into training and testing balanced datasets to deal with the class imbalance problem [11, 45]. Another two studies applied class weighting techniques, which assign higher weights to instances of the minority class during training [21, 29]. For three studies, methods to address class imbalance were unclear and only described addressing class imbalance using "10-fold cross" validation [26], jackknife or "leave one out" procedure [23], and stratified random sampling [43].

## 3.8 Missing data

Missing data is a common challenge in research studies and can have a significant impact on the results and conclusions of the study. Despite its importance, only 20 studies in this scoping review reported how they handled missing data in their studies. Among the identified studies, the most common approach was the removal of rows with missing data, which was used by 12 studies [21–25, 28, 38, 40, 53, 55, 58]. This approach excludes observations with missing data values before training the AI model. However, this approach not only limits the data used in developing the predictive models but also poses a risk of producing biased results as the sample of the target population may not be adequately representative [4, 5]. The other eight studies used imputation methods to fill in missing data, with seven studies using various imputation methods, such as KNN imputation [30], gaussian imputation [54], CART imputation for continuous variables [41], a nonparametric imputation method based on random forest [39], and imputation using the arithmetical mean of the corresponding variable [52]. Only one study [57] compared different methods, including missing data removal and mean, median, and mode imputation, while another study [32] did not perform anything for missing data.

## 3.9 Feature ranking and explainability

Feature ranking and explainability are essential aspects of any predictive model as they can provide insight into which factors are most important in driving the prediction. In the context of T2DM precision medicine, feature ranking and explainability can help understand which risk factors and biomarkers are most influential in determining disease progression. This can help healthcare professionals to gain insights into which areas to target for intervention and prevention. Additionally, interpretable models can increase healthcare professionals' and patients' trust and acceptance of AI-based predictive tools.

In this scoping review, various feature ranking and explainability techniques were reported in almost half of the included studies. Some studies employed permutation feature importance [21, 31, 53, 57], while others used the feature importance built-in functions of algorithms such as decision trees [41, 43, 44], XGBoost [28, 34], RF [11, 48], and HMM [23]. LR was used in three studies [29, 39, 40] to determine feature importance, and two studies ranked the relative importance of risk factors based on their contribution to variance [22, 47]. The significance of risk factors was also explained using Shapley plots in four studies [28, 31, 32, 34]. In one study utilizing DL [35], the model was interpreted using the integrated gradient algorithm to identify the most critical areas in the image.

## 3.10 Risk factors and biomarkers

Half of the studies (n=21) reported the final risk predictors identified by their AI models. Table 4 summarizes the reported biomarkers and risk factors across unimodal and multimodal AI methods utilized in the studies included in this scoping review.

**Table 4** List of reported biomarkers and risk factors by unimodal and multimodal AI models

| Data modality | Biomarker or risk factor category | Biomarker or risk factor | Number of studies | Study References |
|---|---|---|---|---|
| EHR | Anthropometric measures | BMI | n=13 | [11, 22–24, 28, 33, 34, 39, 40, 43, 44, 47, 57] |
| | | Waist circumference (WC) | n=3 | [11, 28, 44] |
| | Biochemical markers | fasting plasma glucose (FPG) | n=11 | [11, 23, 28, 33, 34, 40, 41, 43, 44, 47, 57] |
| | | glycated hemoglobin (HbA1c) | n=5 | [11, 29, 33, 40, 57] |
| | | triglycerides (TG) | n=11 | [11, 23, 28, 29, 31, 33, 40, 43, 44, 47, 57] |
| | | high-density lipoprotein (HDL) | n=5 | [11, 23, 31, 34, 40] |
| | | low-density lipoprotein (LDL) | n=2 | [34, 40] |
| | | alanine transaminase (ALT) | n=5 | [28, 29, 31, 34, 47] |
| | | aspartate transaminase (AST) | n=2 | [31, 47] |
| | | Total cholesterol (TC) | n=3 | [28, 29, 40] |
| | | Gamma-glutamyl transferase (GGT) | n=2 | [47, 57] |
| | Sociodemographic data | age | n=9 | [21–24, 33, 34, 40, 44, 57] |
| | | sex | n=4 | [24, 44, 47, 57] |
| | Medical history | family history of diabetes (FHD) | n=5 | [22, 24, 39, 43, 57] |
| | Lifestyle factors | alcohol intake | n=3 | [22, 47, 57] |
| | Blood pressure | hypertension | n=4 | [22–24, 29, 40, 47] |
| | Other risk factors and biomarkers* | Other risk predictors* | n=1 | |
| Multi-omics | Metabolomic biomarkers | novel markers (α-tocopherol, [Hyp3]-BK, X-12063 and X-13435) and known markers (glucose, mannose and α-HB) | n=1 | [26] |
| | | PC ae C40:5 and SM (OH) C14:1 | | [50] |
| | | riboflavin, cnidioside A, 2-methoxy-5-(1H-1, 2, 4-triazol5-yl)- 4-(trifuoromethyl) pyridine, 7-methylxanthine, and mestranol | n=1 | [48] |
| Imaging | Retinal biomarkers | Retinal biomarkers such as such as vascular tortuosity, venous dilatation, retinal haemorrhage and cotton wool spots | n=1 | [35] |

* Other risk predictors appearing only once in the final model: 2-hour postprandial plasma glucose (2h-PCPG) [26], waist-to-height ratio (WHtR) [26], family history of hypertension [24], smoking [57], physical activity [57], income [22], health insurance [22], occupation [43], chronic liver disease [29], dyslipidemia [22], hypertension [22–24, 29], cardiovascular disease [22], obstructive sleep apnea [29], hypersomnia with sleep apnea [29], hyperlipidemia [29], anemia [29], impaired fasting glucose [29], acute bronchitis [29], abnormal blood chemistry [29], medications (Metformin [29], antiarthritics [29], nonsteroidal anti-inflammatory drugs [29]), serum albumin [24], serum uric acid [57], serum aldosterone [11], serum leptin [11], hematocrit [24], urea [24], health insurance [22], blood glucose level (BGL) [39], left ventricular mass [11], Mean Arterial Pressure (MAP) [43], sodium [24], Inflammatory Markers (high Sensitivity C-Reactive Protein (hs-CRP) [21], log(hsCRP) [21], fibrinogen [21]), Homeostatic Model Assessment of Insulin Resistance (HOMA-IR) [21], Change in Glucose Level From 120 to 60 Minutes After a Meal (G120-60) [53], Change in Glucose Level From 30 to 0 Minutes After a Meal (G30-0 ) [53], Area Under the Glucose Curve From 0 to 120 Minutes After a Meal (AuG0-120 ) [53].

The final reported risk factors and biomarkers varied across the studies. EHR-based biomarkers such as anthropometric measures, glycemic traits, blood lipids, sociodemographic data, and liver enzymes were commonly reported. Additionally, metabolomic biomarkers, imaging-based biomarkers, and other blood biomarkers were reported. Unimodal models, particularly those based on EHRs, reported final risk predictors that most of them were consistent with previous literature and well-understood in terms of their relationship with the underlying biology. These studies' most commonly reported predictors of T2DM progression were BMI, FPG, TG, age, and FHD. Furthermore, liver enzyme biomarkers such as ALT, and AST were investigated and reported by five and three studies, respectively. However, other biochemical markers associated with T2DM risk were rarely investigated and reported with regard to disease prediction, including inflammatory biomarkers (hs-CRP), and fibrinogen), plasma adiponectin, Leptin, Albumin, and aldosterone.

On the other hand, only four studies that used multimodal models reported the final predictors and were able to identify new metabolomic and imaging biomarkers, that unimodal models did not identify. Three studies reported metabolomic biomarkers, some of which were known and others were novel, and one study reported retina scan-based biomarkers. In terms of metabolomic biomarkers, a study [26] found novel markers associated with T2DM progression, such as -tocopherol, [Hyp3]-BK, X-12063, and X-13435, as well as known markers like glucose and mannose. Another study [48] identified five newly discovered metabolic markers, including riboflavin, cnidioside A, 2-methoxy5-(1H-1, 2, 4-triazol-5-yl)-4-(trifluoromethyl) pyridine, 7-methylxanthine, and mestranol. Another study [49] presented a metabolomic signature that provides valuable insight into the etiology of the transition to T2DM in women with previous gestational diabetes mellitus (GDM). The study's prediction revealed that two novel metabolites, PC ae C40:5 and SM (OH) C14:1, are predictive of incident T2DM. In terms of imaging-based biomarkers, one study [35] identified retinal biomarkers associated with the development of T2DM, including vascular tortuosity, venous dilatation, retinal haemorrhage, and cotton wool spots, which ophthalmologists commonly use to diagnose retinal diseases.

Some multimodal models combined polygenic risk scores with metabolomic markers [26, 50]. Others combined retina images with clinical risk factors and reported that this combination improved the model performance in discrimination [35]. However, these studies have not fully reported all the important biomarkers of the different modalities. They have not fully shown the interaction and contribution of the other modalities in the final prediction. This lack of transparency in the prediction process can make it difficult to understand how these multimodal models make predictions and decisions, which can be a barrier to their adoption in clinical practice [62]. Furthermore, the interpretability and complexity of such multimodal models can make it challenging to understand the underlying biology and the interactions between different modalities, which highlights the need for further research to understand the contribution of these multimodal models to identify important predictive biomarkers and enable precision prevention.

### 3.11 Validation approaches

The next crucial step after model development is to evaluate model performance. There are two ways to evaluate performance: (1) internal validation, which involves evaluating the model's performance on the same dataset used to train it like split sampling or cross-validation techniques; and (2) external validation, which involves using a different data set entirely. In the current review, almost all studies reported internal validation (n=38), while a smaller proportion (n=5) used external validation to assess their predictive models generalizability using external independent datasets [26, 34–36, 58].

When validating a prediction model's performance internally, the holdout method is not a suitable utilization of the available data as it lessens the sample size [63, 64]. However, it was the most commonly used method for internal validation in the studies (n=18)[11, 20, 28, 29, 33–35, 37–43, 47, 49, 55, 56]. Cross-validation techniques are preferred as they use all the data for model development and validation[63, 64]. K-fold cross-validation, in which the data are partitioned into k equally sized folds, and the model is trained and evaluated k times, each time using a different fold as the test set, was used in 17 studies (n = 17) [21, 25–27, 30, 31, 44, 46, 48, 50–53, 57]. In contrast, leave-one-out cross-validation was used in one study (n = 1) [23]. The train-valid-test technique was also used in three studies (n = 3) [22, 32, 45], in which the dataset was partitioned into three subsets: a training set, a validation set, and a test set. One study [58] did not mention the type of validation approach used.

## 3.12 Predictive performance

The evaluation metrics used to assess the predictive performance of AI-based T2DM risk prediction models are highly diverse. Most studies utilized discrimination metrics, particularly the AUC (n=32), but did not provide information on model calibration. Standard classification measures were also reported in almost half of the studies, including accuracy, specificity, precision, sensitivity, and F1 score. Other metrics, such as the Youden index, net reclassification improvement (NRI), integrated discrimination improvement (IDI), root mean squared error (RMSE), Jaccard similarity, positive predictive value (PPV), negative predictive value (NPV), Cohen's Kappa, geometric mean (G-Mean), and Matthews Correlation Coefficient (MCC), were also utilized. While calibrating AI models is crucial for predictive performance assessment, only a few studies, five in total, evaluated their models' calibration using measures such as the Brier score, calibration plot, and Hosmer-Lemeshow test. The distribution of the most common reported performance measures used across the included studies is presented in Table 5.

**Table 5** The distribution of evaluation metrics in the included studies.

| Performance measures | Number of studies | References |
| --- | --- | --- |
| AUC | n=32 | [11, 20–37, 39–41, 45–50, 52, 54–56] |
| Accuracy | n=15 | [11, 21, 27, 28, 30, 31, 36, 42–44, 50–53, 57] |
| Specificity | n =15 | [11, 21, 27, 28, 30, 41–43, 47, 48, 51–55] |
| Sensitivity | n = 20 | [11, 21, 27, 28, 30, 31, 35, 36, 41–43, 47, 48, 50–55, 57] |
| Precision | n = 7 | [27, 31, 36, 42, 43, 50, 57] |
| F1 score | n= 8 | [28, 31, 41, 46, 50, 52, 55, 57] |
| Positive Predictive Value | n= 5 | [21, 41, 47, 55, 57] |
| Negative Predictive Value | n= 5 | [21, 41, 47, 55, 57] |
| Net reclassification improvement (NRI) | n= 4 | [37, 45, 46, 49] |
| Integrated discrimination improvement (IDI) | n= 3 | [26, 46, 49] |
| Brier score | n=2 | [37, 46] |
| calibration plot | n=2 | [32, 34] |

| | | | |
|---|---|---|---|
| Hosmer-Lemeshow test | n= 1 | | [47] |

## 3.13 Data resources

The included studies in this scoping review utilized various datasets to develop and evaluate AI models for T2DM risk prediction. These datasets were sourced from both private and public sources. Private datasets were obtained from various sources, including hospitals, clinics, and research institutions, and are not available to the general public. Public datasets, on the other hand, are made available to the general public and these datasets are often collected by government agencies or research institutions and are intended to be used for research purposes. Out of the 40 studies included in this scoping review, 22 utilized publicly accessible datasets [22, 23, 26, 30, 31, 33, 34, 38, 40– 45, 49**?** – 55], while 18 studies relied on private data sources (data that are not publicly available) [11, 20, 21, 24, 25, 27–29, 32, 35–37, 39, 46, 48, 56–58] . It is worth noting that almost all of the public datasets are not freely accessible due to containing sensitive personal health information. Access to the datasets is granted on a case-by-case basis, subject to approval by the study's Data and Research Committee. Therefore, researchers may need to seek approval or permission for access, and some may also require payment of access fees. The most commonly reported public datasets in the included studies were the
Tehran Lipid and Glucose Study (TLGS), Canadian Primary Care Sentinel Surveillance Network (CPCSSN), and San Antonio Heart Study (SAHS), with each being used in three studies. Other datasets were only reported once per study. Appendix 2 in the supplementary information provides further details on these datasets, and Table 6 summarizes the most commonly reported datasets used in the included studies. For further details on the other datasets used in the studies included in this scoping review, interested readers can refer to the papers cited in the references.

**Table 6** The most commonly used public datasets.

| Dataset name | Description | Population | Da | Study reference |
|---|---|---|---|---|
| SAHS | It is a longitudinal epidemiological investigation that seeks to explore the incidence of T2DM and cardiovascular disease in the population of San Antonio, involving both Mexican Americans and nonHispanic whites. | USA | https://www.ncbi. nlm.nih.gov/projects/ gap/cgi-bin/study.cgi? study id=phs001215. v3.p2 | [23, 33, 40] |
| CPCSSN | It is a primary care electronic medical record surveillance system that collects data from participating primary care clinics in Canada for various health research purposes, including the development and evaluation of chronic disease management programs, monitoring disease trends, and improving primary care services | Canada | http://cpcssn.ca/join-cpcssn/for-researchers/ | [51–53] |
| TLGS | It is a large-scale, longitudinal, population-based study conducted in Tehran, Iran. The study was initiated in 1999 with the aim of assessing the prevalence and incidence of non-communicable diseases and their risk factors among Iranian adults. | Iran | https://endocrine.ac.ir/page/Tehran-Lipid-and-Glucose-Study-TLGS?lang=en | [41–43] |

### 3.14 Reproducibility

This review also summarized the studies in which the implementation code has been made available. Out of the studies analyzed, only a small number, four to be precise, made their code publicly accessible [31, 38, 40, 46]. This highlights the need for authors to make their code publicly available to facilitate reproducibility.

## 4 Discussion

This scoping review has comprehensively analyzed the current state of AI-based models for T2DM risk prediction in the published literature. This section summarizes the key findings and outlines potential future directions for research in this area.

A total of 40 studies were included, and the results showed promising performance for AI models for T2DM risk prediction. Many data modalities and modeling techniques were used to develop these prediction models. EHR data was the most common data type used in the studies. This data is often used alone or in combination with other modalities, such as multi-omics and imaging data. Multi-omics data, including genomics and metabolomics, were the second most used data modality, while imaging data such as CT scans and retinal scans were the least used data.

The majority of studies examined the use of unimodal AI models for predicting the risk of T2DM. These studies used different AI algorithms to develop predictive models, with classical ML models being the most widely used such as tree-type (DT and RF), SVM, KNN, and ensemble learning models. Unimodal models have shown moderate to high performance with an average AUC of 0.809. However, it is important to note that it is difficult to determine which AI model is performing the best, as the type and combination of input risk predictors, and, to some extent, prediction horizon can significantly influence the model's performance. For example, the XGBoost algorithm was used in three unimodal studies [28, 32, 34], and the reported AUC was 0.91,0.826,0.67 respectively. Each of these studies used different datasets with different sample sizes and combinations of risk predictors, making it perform well with certain combinations of risk factors but not as well with others. Moreover, the prediction horizon duration also significantly impacts the prediction performance of the same model. In studies that performed prediction over different time periods [22, 24, 47, 54], we observed clearly that the discriminatory power decreased as the prediction horizon increased. Unimodal AI models, while useful, may not capture the complexity of T2DM risk prediction as the individual's state is characterized by a spectrum of modalities, ranging from EHR, multi-omics to imaging. Such single-modality models neglect the broader clinical context, which inevitably diminishes their potential. On the other hand, multimodal models have the advantage of incorporating multiple data sources and modalities, providing a more holistic view of the individual and potentially improving the prediction accuracy of T2DM risk.

In this scoping review, multimodal prognostic models for T2DM were represented by a smaller proportion of the studies included. The most commonly used data modalities for such models were multi-omics data in combination with EHR data, followed by imaging data with EHR metadata. None of the studies included in this review have combined the three multi-omics, imaging, and EHR-based risk factors into one predictive model. These models employ two commonly used fusion strategies, early fusion and joint fusion. Early fusion was the dominant approach for multimodal learning and was commonly used to combine multi-omics data with EHR data. On the other hand, joint fusion, which combines multiple data modalities at the feature level, was less commonly used and primarily used to integrate imaging with EHR data. In terms of predictive performance, the results of this scoping review showed that the prediction performance of these multimodal models was generally better than that of unimodal models, with an average AUC value of 0.89. Furthermore, many of the multimodal studies compared their results to those of their unimodal counterparts and consistently showed a performance improvement when leveraging multimodal data [26, 35–37, 45, 46, 48, 49]. This result suggests that incorporating multimodal models can be beneficial in precision diabetes medicine. This finding aligns with previous reviews on cancer research [62] and cardiovascular disease care [65]. One of the main advantages of multimodal AI models is their ability to identify complex interactions between different risk factors, which may not be apparent when using a single data modality. Their use can lead to more accurate risk predictions, enabling personalized prevention and management strategies to be developed for individuals at high risk for T2DM.

It can be reasonably stated that the multimodal approach enhances the predictive performance compared to single-modal methods. This supports the idea that multimodal AI may be a valuable tool in precision medicine for further subgrouping patients and identifying their unique health fingerprints, enabling more precise preventive measures targeting [60]. However, it is also important to note that the development of multimodal models also poses certain challenges, such as the time-consuming nature of their development, data concatenation, and lower scalability [59]. Additionally, the increased complexity of these models, which often incorporate multiple data sources, makes it difficult to understand the interaction between the different modalities and how the model makes its predictions. This lack of interpretability may impede their implementation in clinical settings and be a challenge for clinicians and researchers who need to understand the underlying mechanisms and reasoning behind the models' predictions in order to trust and use them in clinical practice. Given the limited number of multimodal AI models for T2DM currently available, further research is needed to investigate the use of multimodal AI models and to determine the best strategies for fusing different data modalities. Furthermore, it is of utmost importance to focus on interpretability and explainability when developing these models, as this will facilitate their successful implementation in clinical settings.

Despite the promising results of AI models for T2DM prognosis, it is important to note that the quality and completeness of performance reporting among the studies included in this scoping review were variable. Adequate information on various dimensions of predictive performance, such as discrimination and calibration, is crucial in determining the value of a prediction model. However, only a limited number of studies reported calibration measures, and many reported only a single dimension of performance, such as the AUC or classification measures. Uncalibrated models may have limited applicability in practical, real-world situations [66]. The lack of detailed performance reporting in the studies presents a challenge in determining the generalizability and practicality of these AI models in real-world settings. Thus, we recommend that future studies should emphasize comprehensively reporting predictive performance, including metrics such as discrimination, calibration, and classification measures. This will aid in evaluating the validity and usefulness of AI-based T2DM prognostic models for precision medicine.

Despite the advancements in the development of AI-based prediction models for T2DM prognosis, this scoping review identifies some barriers that hinder the progress of knowledge and their clinical usability. The first barrier is that the majority of studies in the literature relied on hold-out internal validation approaches. The performance estimate using this approach can be highly dependent on which data points end up in the training set and which end up in the validation set, which can lead to high variance in the performance estimate.

This can make it difficult to assess the true performance of the model. Moreover, this approach presents a concern as it reduces the sample size available for model development and may not utilize the data effectively [63, 64]. Furthermore, only five studies performed external validation raising concerns about their generalizability and hindering their practical application. This finding aligns with previous reviews that noted a lack of external validation for prediction models [67]. There is growing evidence that many areas of scientific research are experiencing a replicability crisis, such as precision psychiatry [9], genetic behavior research [68], and cancer research [69]. Therefore, we recommend providing external validation and testing of AI models in different settings and populations to establish a solid foundation for their implementation in clinical practice and improve their potential to guide prevention strategies for T2DM. This will aid in addressing one of the largest barriers for precision prognostics in real-world implementations. In Fig. 4, we highlighted the limitations of the included studies that occur in the data processing, model development, and evaluation.

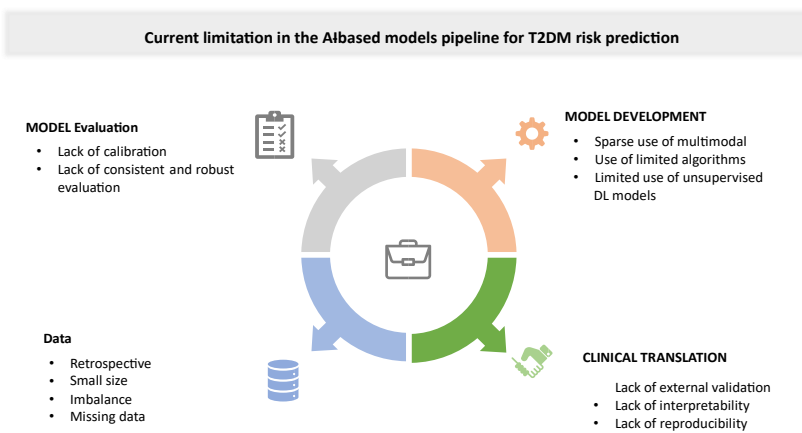

**Fig. 4** Key limitations and Challenges in the field of AI models for T2DM prognosis.

ML is inherently iterative; thus, the optimal predictive model could be generated by comparing a combination of algorithms [16]. However, the included studies did not frequently adopt such extensive modeling approaches. For instance, only half of the studies compared multiple algorithms and selected the best one, while the other studies were developed using a single algorithm.

Unsupervised ML, which has the ability to reveal the inherent structures and patterns within multidimensional data [70], was rarely utilized. Therefore, we recommend that future research consider using multiple AI models to identify the most precise one.

Additionally, interpretability and explainability methods, such as permutation importance or Shapley values, can provide insights into which variables had the most significant influence on the prediction. The capability of AI techniques to learn abstract feature representations raises concerns about the possibility of the models relying on fake shortcuts for predictions rather than learning clinically relevant information; this may result in models that are unable to generalize effectively when faced with new data or exhibit discriminatory behavior towards certain populations [27, 28]. On the other hand, models can discover novel and clinically relevant biomarkers that can enable precision medicine, allowing for understanding specific risk factors associated with the development of disease and enabling clinicians to match patients with similar complication risks and optimize therapies. However, many studies in this review did not interpret the predictions made by their models and did not report the crucial risk predictors. They relied solely on performance metrics to indicate the high performance of their models. It is of utmost importance to comprehend the reasoning behind a model's predictions, particularly in clinical applications. The objective of clinical machine learning studies is not only to make predictions but also to gain valuable insights. As a result, there has been a shift in focus from solely evaluating prediction performance to a greater emphasis on comprehending algorithm dynamics. This has become a priority in research and has led to a surge in the number of related studies conducted [71, 72].

Identifying important predictors in T2DM prediction models is essential for precision medicine and prevention as it helps target interventions for highrisk individuals and aids in developing new treatments. However, around half of the studies examined the final risk factors and biomarkers associated with T2DM risk, while the other half did not report them. The studies included in this review identified a wide range of biomarkers and risk factors associated with the development of T2DM. These included traditional risk factors, such as BMI, blood cholesterol measurements, FPG, age, FHD, and HbA1c, which were consistent with prior research findings. With the advancement of molecular biology and medical imaging, several molecular markers, such as gene expression, metabolomic, and imaging markers, have become potential predictors for T2DM. Only a few studies that used multimodal models reported the final predictors and were able to identify new metabolomic and imaging biomarkers not identified by unimodal models. However, these studies did not comprehensively report the different biomarkers of the various modalities and did not sufficiently demonstrate the interaction and contribution of the diverse modalities to the final prediction. Therefore, it is recommended that future multimodal studies in this field provide a more detailed examination of the specific biomarkers of the various combined modalities used in their models, as well as a more thorough examination of the interactions and contributions of these different factors to the final prediction. This would help to improve the interpretability and clinical utility of these models and aid in their implementation in real-world settings.

Additionally, this review highlights a number of methodological flaws raised in the studies that hinder the implementation of AI in clinical settings and precision medicine. These limitations include small sample sizes, retrospective data, imbalanced samples, inadequate handling of missing data, and challenges in interpreting the models employed. Small sample sizes often result in poor model fitting and generalizability, with some studies having as few as 244 participants and fewer than 1000 in six studies. While some studies used large cohorts, many others did not. Furthermore, there was a limited number of investigations based on prospective data, with most models developed retrospectively from research datasets assembled for other purposes. The neglect of sample imbalances often leads to biased models and misleading performance metrics [73]. Additionally, the inadequate handling of missing data can skew the results, and a comparison of different imputation methods should be part of the reporting process [73].

It is important to note that reproducibility is a key aspect of scientific research, and the development of AI models for T2DM prognosis is no exception. The availability of model code and data is an important aspect of reproducibility as it allows other researchers to verify and build upon the work independently. In this scoping review, we found that the majority of the included studies did not report the availability of their model's code or data, making it difficult to reproduce their findings. This lack of transparency in the reporting of model code and data availability can impede the replication and validation of the models, which in turn hinders the progress of research in this field. Due to privacy regulations such as the "health insurance portability and accountability act" (HIPAA), sharing medical data may not be feasible. However, other measures to promote transparency can be taken, such as requiring authors to provide a summary of their data sample and statistical information about the complete dataset, including the number of data points, key variables, distribution, and class information. A more ideal solution would be to create a synthetic dataset derived from the original data [74–76]. Future studies in the field of AI models for T2DM prognosis are recommended to report the availability of model codes and data to ensure that their findings can be independently verified and replicated by other researchers. This would not only improve the reproducibility of the research but also facilitate the validation and implementation of the models in real-world settings.

## 4.1 Limitations

In conducting this scoping review, several limitations should be acknowledged. Firstly, the literature search was restricted to English-language articles and excluded grey literature, which may result in some studies being omitted.

Nevertheless, it is believed that including additional articles in the review would not have significantly impacted the findings. Secondly, the inclusion criteria for this review were narrow, only including studies that specifically evaluated the use of AI models for T2DM risk prediction. As a result, this review does not provide a comprehensive overview of all the research on T2DM and AI. Additionally, as this study's focus was to conduct a scientific paper profiling of AI models for T2DM prognosis, a thorough evaluation of the individual methodological quality of the included studies was not conducted. However, insights were offered on the potential limitations in methodology that may have influenced the results. Because positive results are typically reported disproportionately, publication bias might be another limitation of this review. This bias may result in overestimating the benefits associated with AI-based models in risk prediction. There is significant heterogeneity among the studies included in this review in terms of the data sources, study populations, and evaluation metrics, making it difficult to compare the results of different studies directly. Finally, this scoping review only covers the current state of the use of AI models for T2DM prognosis and does not provide a comprehensive evaluation of their potential benefits and limitations. Thus, future studies are needed to further evaluate the feasibility, accuracy, and potential benefits of using AI models for T2DM prognosis.

# 5  Conclusion

Precision medicine and personalized prevention strategies aim to tailor treatment and prevention plans to an individual's unique characteristics, to improve health outcomes, and reduce the risk of developing diseases such as T2DM. AI models can explore complex and diverse data to identify prognostic biomarkers, which allows for the identification of individuals at high risk for developing T2DM, enabling earlier intervention and prevention efforts. Our review found a growing body of literature on using AI predictive models, both unimodal and multimodal, for T2DM prognosis. The results of this review suggest that AI models have promising potential in predicting the future development of T2DM. While unimodal models have shown varied performance, multimodal models demonstrated improved performance compared to their unimodal counterparts. However, some several challenges and considerations need to be addressed to realize this potential. Additionally, as with any significant medical advancement, there is a need for thorough validation and evaluation through clinical trials and prospective studies to verify the potential benefits claimed by AI models. The role of AI in medicine is not autonomous but rather a partnership between AI models and human expertise that will drive progress in the field. Despite limitations and challenges, it is our responsibility to capitalize on the benefits of AI methods to accelerate the discovery and translation of advances into clinical practice for the benefit of patients and healthcare providers [77].

## Declarations

- Conflict of interest/Competing interests: The authors declare no competing interests.
- Availability of data and materials: The data generated during this scoping review is provided as supplementary materials.
- Authors' contributions: F.M. and Z.S. contributed to conceptualization. F.M. administered the project. F.M. and H.R.A. performed screening, study selection, and data extraction. F.M. performed data synthesis and contributed to writing—the original draft. Z.S., N.E., and N.Y. performed writing—review, and editing. Z.S. supervised the study. All authors read and approved the final manuscript.